# Otem&Utem: Over- and Under-Translation Evaluation Metric for NMT

Jing Yang[1,2], Biao Zhang[1], Yue Qin[1], Xiangwen Zhang[1], Qian Lin[1], and Jinsong Su[*1,2]

[1]Xiamen University, Xiamen, China
[2]Provincial Key Laboratory for Computer Information Processing Technology,
Soochow University, Suzhou, China
{jingy, zb, qinyue, xwzhang, linqian17}@stu.xmu.edu.cn
jssu@xmu.edu.cn

July 24, 2018

**Abstract**

Although neural machine translation(NMT) yields promising translation performance, it unfortunately suffers from over- and under-translation issues [Tu *et al.*, 2016], of which studies have become research hotspots in NMT. At present, these studies mainly apply the dominant automatic evaluation metrics, such as BLEU, to evaluate the overall translation quality with respect to both adequacy and fluency. However, they are unable to accurately measure the ability of NMT systems in dealing with the above-mentioned issues. In this paper, we propose two quantitative metrics, the *Otem* and *Utem*, to automatically evaluate the system performance in terms of over- and under-translation respectively. Both metrics are based on the proportion of mismatched n-grams between gold reference and system translation. We evaluate both metrics by comparing their scores with human evaluations, where the values of Pearson Correlation Coefficient reveal their strong correlation. Moreover, in-depth analyses on various translation systems indicate some inconsistency between BLEU and our proposed metrics, highlighting the necessity and significance of our metrics.

***Keywords***— Evaluation Metric Neural Machine Translation Over-transaltion Under-transaltion.

## 1 Introduction

With the rapid development of deep learning, the studies of machine translation have evolved from statistical machine translation (SMT) to neural machine translation (NMT) [Sundermeyer *et al.*, 2014; Sutskever *et al.*, 2014]. Particularly, the introduction of attention mechanism [Bahdanau *et al.*, 2015] enables NMT to significantly outperform SMT. By now, attentional NMT has dominated the field of machine translation and continues to develop, pushing the boundary of translation performance.

---

*Corresponding author.

Despite of its significant improvement in the translation quality, NMT tends to suffer from two specific problems [Tu *et al.*, 2016]:

- *Over-translation* where some words are unnecessarily translated for multiple times, and

- *Under-translation* where some words are mistakenly untranslated.

To address these issues, researchers often learn from the successful experience of SMT to improve NMT [Tu *et al.*, 2016; Cohn *et al.*, 2016; Feng *et al.*, 2016; Yang *et al.*, 2017]. In these studies, the common practice is to use the typically used translation metrics, such as BLEU [Papineni *et al.*, 2002], METEOR [Banerjee and Lavie, 2005] and so on, to judge whether the proposed models are effective. However, these metrics are mainly used to measure how faithful the candidate translation is to its gold reference in general, but not for any specific aspects. To large extent, they are incapable of accurately reflecting the performance of NMT models in addressing the drawbacks mentioned previously.

Let us consider the following example:

- **Source Sentence**: *tā hūyù měiguó duì zhōngdōng hépíng yào yǒu míngquè de kànfǎ , bìng wèi cǐ fāhuī zuòyòng , yǐ shǐ liánhéguó yǒuguān juéyì néng dédào qièshí zhíxíng .*

- **Reference 1**: *he urged that the united states maintain a clear notion of the peace in the middle east and play its due role in this so that the un resolutions can be actually implemented .*

- **Reference 2**: *he urged u.s. to adopt a clear position in the middle east peace process and play its role accordingly . this is necessary for a realistic execution of united nations ' resolutions .*

- **Reference 3**: *he called for us to make clear its views on mideast peace and play its role to ensure related us resolutions be enforced .*

- **Reference 4**: *he called on the us to have a clear cut opinion on the middle east peace , and play an important role on it and bring concrete implementation of relative un resolutions .*

- **Candidate 1**: *he called on the united states to have a clear view on peace in the middle east peace and play a role in this regard so that the relevant un resolutions can be effectively implemented .* (**BLEU = 45.83**)

- **Candidate 2**: *he called on the united states to have a clear view on in the middle east and play a role in this regard so that the relevant un resolutions can be effectively implemented .* (**BLEU = 46.33**)

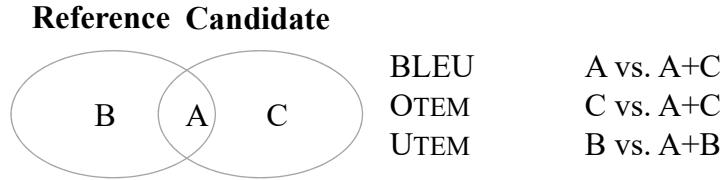

Figure 1: Intuitive comparison of BLEU, OTEM and UTEM. We use gray circle to illustrate the gold reference (left) and candidate translation (right). Capital "A" denotes the matched n-grams, while capital "B" and "C" denotes the mismatched parts.

Obviously, two candidate translations have different translation errors. Specifically, in Candidate 1, the Chinese word "*hépíng*" is over-translated, where "*peace*" appears twice in Candidate 1. In contrast, in Candidate 2, "*hépíng*" is under-translated, for its translation is completely omitted. However, the BLEU metric is unable to distinguish between these two kinds of translation errors and assigns similar scores to these two candidates. This result strongly indicates the incapability of BLEU in detecting the over- and under-translation phenomena. Therefore, it is significant for NMT to explore better translation quality metric specific to over-translation and under-translation.

In this paper, we propose two novel automatic evaluation metrics: "OTEM" short for *over-translation evaluation metric* and "UTEM" short for *under-translation evaluation metric*, to assess the abilities of NMT models in dealing with over-translation and under-translation, respectively. Both metrics count the lexical differences between gold reference and system translation, and provide quantitative measurement according to the proportion of mismatched n-grams. Figure 1 shows the intuitive comparison among BLEU, OTEM and UTEM. The BLEU calculates the precision of matched n-grams ($A$) over the whole reference ($A + C$). By contrast, the OTEM focuses on the proportion of repeated n-grams in the candidate translation ($C$) over the whole candidate ($A + C$), and the UTEM estimates the proportion of untranslated n-grams in the reference ($B$) over the whole reference ($A + B$). Clearly, BLEU is correlated with both UTEM and OTEM but incapable of inferring them.

To evaluate the effectiveness of our proposed metrics, we conducted translation experiments on Chinese-English translation using various SMT and NMT systems. We draw the following two conclusions: (1) There exists strong correlations between the proposed metrics and human evaluation measured by the Pearson Correlation Coefficient, and (2) The significant improvement in terms of BLEU score doesn't imply the same improvement in OTEM and UTEM, by contrast, our proposed metrics can be used as supplements to the BLEU score. Moreover, further analysis shows the diverse characteristics of the NMT systems based on different architectures.[1]

---

[1] Evaluation scripts are available at https://github.com/DeepLearnXMU/Otem-Utem.

## 2  Related Work

Usually, most of the widely-used automatic evaluation metrics are used to perform the overall evaluation of translation quality. On the whole, these metrics can be divided into the following three categories: (1) *The Lexicon-based Metrics* are good at capturing the lexicon or phrase level information but can not adequately reflect the syntax and semantic similarity [Leusch *et al.*, 2003; Nießen *et al.*, 2000; Papineni *et al.*, 2002; Doddington, 2002; Lin, 2004; Snover *et al.*, 2006; Banerjee and Lavie, 2005; Chen and Kuhn, 2011; Chen *et al.*, 2012; Han, 2017]; (2) *The Syntax/Semantic-based Metrics* exploit the syntax and semantic similarity to evaluate translation quality, but still suffer from the syntax/semantic parsing of the potentially noisy machine translations [Liu and Gildea, 2005; Owczarzak *et al.*, 2007; Giménez and Màrquez, 2007; Mehay and Brew, 2007; Lo and Wu, 2011; Lo *et al.*, 2012; Lo and Wu, 2013; Yu *et al.*, 2014]; (3) *The Neural Network-based Metrics* mainly leverage the embeddings of the candidate and its references to evaluate the candidate quality [Chen and Guo, 2015; Guzmán *et al.*, 2015; Gupta *et al.*, 2015; Han, 2017].

Since our metrics involve n-gram matching, we further discuss the two subclasses in the first aspect: (1) *Evaluation Metrics based on N-gram Matching*. By utilizing the n-gram precisions between candidate and references, the F-measure, the recall and so on, these metrics attain the goal to evaluate the overall quality of candidate [Papineni *et al.*, 2002; Doddington, 2002; Lin, 2004; Banerjee and Lavie, 2005; Chen and Kuhn, 2011; Chen *et al.*, 2012]. (2) *Evaluation Metrics based on Edit Distance.* The core idea of these metrics [Leusch *et al.*, 2003; Nießen *et al.*, 2000; Snover *et al.*, 2006; Popović and Ney, 2011] is to calculate the edit distance required to modify a candidate into its reference, which can reflect the discrepancy between a candidate and its references.

Our work is significantly different from most of the above-mentioned studies, for we mainly focus on the over- and under-translation issues, rather than measuring the translation quality in terms of adequacy and fluency. The work most closely related to ours is the N-gram Repetition Rate (N-GRR) proposed by Zhang *et al.* [2017], which merely computes the portion of repeated n-grams for over-translation evaluation. Compared with our metrics, the OTEM in particular, N-GRR is much simpler for it completely ignores the n-gram distribution in gold references and doesn't solve length bias problem. To some extent, OTEM can be regarded as a substantial extension of N-GRR.

Meanwhile, the metrics proposed by Popović and Ney [2011] also evaluate the MT translation on different types of errors such as missing words, extra words and morphological errors based on edit distance. However, its core idea extends from WER and PER, and it only takes the word-level information into consideration, while the length bias problem can't be solved similarly. The evaluation of addition and omission can be seen as the simplified 1-gram measurement of OTEM and UTEM theoretically. In addition, Malaviya *et al.* [2018] also presented two metrics to account for over- and under-translation in MT translation. Unlike our model, however, the problem of length bias was also not solved in this work.

## 3 Our Metrics

In this section, we give detailed descriptions of the proposed metrics. The ideal way to assess over-translation or under-translation problems is to semantically compare a source sentence with its candidate translation and record how many times each source word is translated to the target word, which unfortunately is shown to be trivial. Therefore, here we mainly focus on the study of simple but effective automatic evaluation metrics for NMT specific to over- and under-translation.

Usually, a source sentence can be correctly translated into diverse target references which differ in word choices or in word orders even using the same words. Besides that, there are often no other significant differences among the n-gram distributions of these target references. If the occurrence of a certain n-gram in the generated translation is significantly greater than that in all references, we can presume that the generated translation has the defect of over-translation. Similarly, if the opposite happens, we can assume that under-translation occurs in the generated translation. Based on these analyses, we follow Papineni *et al.* [2002] to design OTEM and UTEM on the basis of the lexical matchings between candidate translations and gold references:

$$Otem/Utem := LP * \exp\left(\sum_{n=1}^{N} w_n \log mp_n\right), \quad (1)$$

where $LP$ indicates a factor of length penalty, $N$ is the maximum length of the considered n-grams, and $mp_n$ denotes the proportion of the mismatched n-grams contributing to the metric by the weight $w_n$. It should be noted that here we directly adapt the weight definition of BLEU [Papineni *et al.*, 2002] to ours, leaving more sophisticated definitions to future work. Specifically, we assume that different n-grams share the same contribution to the metric so that $w_n$ is fixed as $\frac{1}{N}$. Although this formulation looks very similar to the BLEU, the definitions of $BP$ and $p_n$, which lie at the core of our metrics, differs significantly from those of BLEU and mainly depend on the specific proposed metrics. We elaborate more on these details in the following subsections.

### 3.1 Otem

As described previously, when over-translation occurs, the candidate translation generally contains many repeated n-grams. To capture this characteristic, we define $mp_n$ to be the proportion of these over-matched n-grams over the whole candidate translation as follows:

$$mp_n = \frac{\sum\limits_{\mathcal{C} \in \{Candidates\}} \sum\limits_{n\text{-}gram \in \mathcal{C}} Count_{over}\left(n\text{-}gram\right)}{\sum\limits_{\mathcal{C}' \in \{Candidates\}} \sum\limits_{n\text{-}gram' \in \mathcal{C}'} Count_{cand}\left(n\text{-}gram\right)}, \quad (2)$$

where $\{Candidates\}$ denotes the candidate translations of a dataset, $Count_{over}(\cdot)$ calculates the over-matched times of the n-gram from the candidate translation, and $Count_{cand}(\cdot)$ records the occurrence of the n-gram in the candidate translation. When referring to $Count_{over}(\cdot)$, we mainly focus on two kinds of over-matched n-grams: (1) the n-gram which occurs in both reference and candidate, and its occurrence in the latter exceeds that in the former; and (2) the n-gram that occurs only in candidate, and its occurrence exceeds 1.

Moreover, we define the over-matched times of *n-gram* as follows:

$$\begin{cases} Count_{cand}(\textit{n-gram}) - Count_{ref}(\textit{n-gram}), & \text{if } Count_{cand}(\textit{n-gram}) > Count_{ref}(\textit{n-gram}) > 0; \\ Count_{cand}(\textit{n-gram}) - 1, & \text{if } Count_{cand}(\textit{n-gram}) > 1 \text{ and } Count_{ref}(\textit{n-gram}) = 0; \\ 0, & \text{otherwise,} \end{cases} \quad (3)$$

where $Count_{ref}(\textit{n-gram})$ denotes the count of *n-gram* in its reference. When multiple references are available, we choose the minimum $Count_{ref}(\textit{n-gram})$ for this function, as we argue that a n-gram is not over-matched as long as it is not over-matched in any reference. Back to the Candidate 1 mentioned in Section 1, $Count_{cand}(\text{``peace''})$ is 2, while $Count_{ref}(\text{``peace''})$ in all references is 1, and thus $Count_{over}(\text{``peace''})$ is calculated as 1.

Another problem with over-translation is that candidates tend to be longer because many unneccessary n-grams are generated repeatedly, which further causes the calculation bias in OTEM. To remedy this, we introduce the length penalty $LP$ to penalize long translations. Formally,

$$LP = \begin{cases} 1, & \text{if } c < r; \\ e^{1-\frac{r}{c}}, & \text{otherwise,} \end{cases} \quad (4)$$

where $c$ and $r$ denote the length of candidate translation and its reference respectively. For multiple references, we select the one whose length is closest to the candidate translation, following Papineni *et al.* [2002].

### 3.2 Utem

Different from OTEM, UTEM assesses the degree of omission in the candidate translation for a source sentence. Whenever under-translation occurs, some n-grams are often missed compared with its reference. Therefore, we define $mp_n$ to be the proportion of these under-matched n-grams over the reference as follows:

$$mp_n = \frac{\sum\limits_{\mathcal{R} \in \{References\}} \sum\limits_{\textit{n-gram} \in \mathcal{R}} Count_{under}(\textit{n-gram})}{\sum\limits_{\mathcal{R}' \in \{References\}} \sum\limits_{\textit{n-gram}' \in \mathcal{R}'} Count_{ref}(\textit{n-gram})}, \quad (5)$$

where $\{References\}$ indicates the gold references from a dataset, and $Count_{ref}(\cdot)$ counts the occurrence of n-gram in the reference.

Note that the above formula only deals with one reference for each source sentence, however, both numerator and denominator in Equation 5 suffer from the selection bias problem, when there are multiple references. In this case, we employ a default strategy to preserve the minimum $Count_{under}(\cdot)$ value as well as the maximum $Count_{ref}(\cdot)$ value for each n-gram based on an optimistic scenario.

As for $Count_{under}(\cdot)$, we mainly consider two types of under-matched n-grams: 1) the n-gram that occurs in both reference and candidate, and its occurrence in the former exceeds that in the latter; and 2) the n-gram that appears only in reference. Furthermore, we calculate their $Count_{under}(\cdot)$ as follows:

$$\begin{cases} Count_{ref}(\textit{n-gram}) - Count_{cand}(\textit{n-gram}), & \text{if } Count_{ref}(\textit{n-gram}) > Count_{cand}(\textit{n-gram}); \\ 0, & \text{otherwise.} \end{cases} \quad (6)$$

In this way, the more parts are omitted in translation, the larger $Count_{under}(\cdot)$ will be, which as expected can reflect the under-translation issue. Still take the Candidate 2 described in Section 1 as an example, we find that $Count_{ref}(\textit{"peace"})$ is 1, so, $Count_{under}(\textit{"peace"})$ is computed as 1.

Furthermore, when some source words or phrases are untranslated, the resulting candidate translation generally tends to be shorter. Accordingly, we also leverage the length penalty $LP$ to penalize short translations, i.e.

$$LP = \begin{cases} 1, & \text{if } c > r; \\ e^{1-\frac{c}{r}}, & \text{otherwise.} \end{cases} \quad (7)$$

where the definitions of $c$ and $r$ are the same as those in Equation 4.

## 4 Experiments

We evaluated our proposed metrics on Chinese-English translation task.

### 4.1 Datasets and Machine Translation Systems

We collected 1.25M LDC sentence pairs with 27.9M Chinese words and 34.5M English words as the training corpus. Besides, we chose the NIST 2005 dataset as the validation set and the NIST 2002, 2003 and 2004 datasets as the test sets. Each source sentence in these datasets is annotated with four different references.

For the sake of efficiency, we only kept the sentences of length within 50 words to train NMT models. In this way, there are 90.12% of parallel sentences were involved in our experiments. As for the data preprocessing, we tokenized Chinese sentences using *Stanford Word Segmenter*[2], and English sentences via *Byte Pair Encoding* (BPE) [Sennrich et al., 2016]. We set the vocabulary size to 30K for NMT model training. For all the out-of-vocabulary words in the corpus, we replaced each of them with a special token UNK. Finally, our vocabularies covered 97.4% Chinese words and 99.3% English words of the corpus.

We carried out experiments using the following state-of-the-art MT systems:

---

[2]https://nlp.stanford.edu/software/segmenter.html

- *PbSMT and HieSMT*: We trained a phrase-based (*PbSMT*) [Koehn et al., 2003] and a hierarchical phrase-based (*HieSMT*) [Chiang, 2007] SMT system using *MOSES* with default settings. We word-aligned the training corpus using *GIZA++* with the option "*grow-diag-final-and*", and trained a 5-gram language models on the GIGAWORD Xinhua corpus using the *SRILM* toolkit[3] with modified *Kneser-Ney* smoothing.

- *RNNSearch*: a re-implementation of the attention-based NMT model [Bahdanau et al., 2015] based on dl4mt tutorial[4]. . We set word embedding size as 620, hidden layer size as 1000, batch size as 80, gradient norm as 1.0, and beam size as 10. All the other settings are the same as in [Bahdanau et al., 2015].

- *Coverage*: an enhanced RNNSearch equipped with a coverage mechanism [Tu et al., 2016]. We used the same model settings as in the above RNNSearch.

- *FairSeq*: a convolutional sequence-to-sequence learning system [Gehring et al., 2017][5]. . We used 15 convolutional encoder and decoder layers with a kernel width of 3, and set all embedding dimensions to 256. Others were kept as default.

- *Transformer*: model [Vaswani et al., 2017] reimplemented by Tsinghua NLP group[6]. We trained the base Transformer using 6 encoder and decoder layers with 8 heads, and set batch size as 128.

### 4.2 Comparison with Human Translation

In theory, our metrics are capable of distinguishing human translation with no over- and under-translation issues from the machine translated ones that may suffer from these issues. To verify this, we collected the translations produced by RNNSearch and one of four references of each source sentence in NIST 2002 dataset. We compare them by calculating the mismatch proportion $mp_n$ against three other gold references. Figure 2 shows the results.

Not surprisingly, with the increase of n-gram length, the proportion of overmatched n-grams drops gradually. This is reasonable because long n-grams are more difficult to be generated repeatedly. By contrast, the proportion of undermatched n-grams grows steadily. The underlying reason is that long n-grams tend to be more difficult to be matched against the reference. No matter how long the n-gram is, our OTEM metric assigns significantly greater scores to the human translations than the machine translated ones. Meanwhile, the scores of our UTEM metric on the human translations are also significantly less than those of machine translation. Besides, it is clear that both OTEM and UTEM metrics

---

[3] http://www.speech.sri.com/projects/srilm/download.html
[4] https://github.com/nyu-dl/dl4mt-tutorial
[5] https://github.com/facebookresearch/fairseq
[6] https://github.com/thumt/THUMT

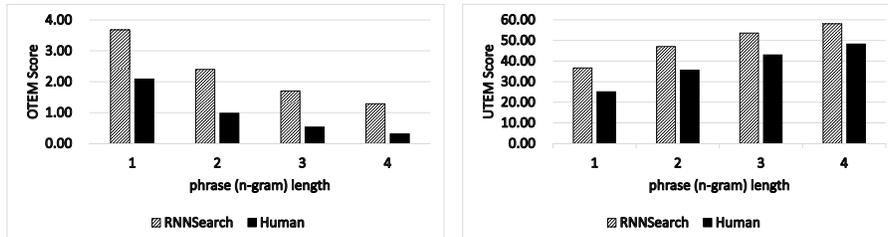

(a) OTEM distinguish human translation from RNNSearch translation

(b) UTEM distinguish human translation from RNNSearch translation

Figure 2: Comparison between RNNSearch and human translation on NIST 2002 dataset in terms of mismatch proportion $mp_n$, where $n$ ranges from 1 to 4. The fig. 2(a) is for OTEM, and the fig. 2(b) is for UTEM.

show great and consistent difference between the evaluation score of RNNSearch and human translation, strongly indicating their ability in differentiating human translations from the machine translated ones.

### 4.3 Human Evaluation

In this section, we investigate the effectiveness of OTEM and UTEM by measuring their correlation and consistency with human evaluation. Existing manual labeled dataset is usually annotated with respect to faithfulness and fluency, rather than over- and under-translation. To fill this gap, we first annotate a problem-specific evaluation dataset. Then we examine our proposed metrics using the Pearson Correlation Coefficient (Pearson's r).

#### 4.3.1 Data Annotation

Following the similar experimental setup in [Papineni *et al.*, 2002], we used the NIST 2002 dataset for this task. In order to avoid selection bias problem, we randomly sampled five groups of source sentences from this dataset. Each group contains 50 sentences paired with candidate translations generated by different NMT systems (including RNNSearch, Coverage, FairSeq and Transformer). In total, this dataset consists of 1250 Chinese-English sentence pairs.

We arranged two annotators to rate translations in each group from 1 (almost no over- or under-translation issue) to 5 (serious over- or under-translation issue) , and average their assigned scores to the candidate translation as the final manually annotated score. The principle of scoring is the ratio of over-tranlated or under-translated word occurrence in candidate translations. It is to be noted that this proportion has a certain subjective influence on scoring according to the length of the candidate and source sentence. For example, with the same over-translated number of words in the candidate(eg. 5 words), the score can change from 2(few words have been over-translated) to 4(a large number of words have

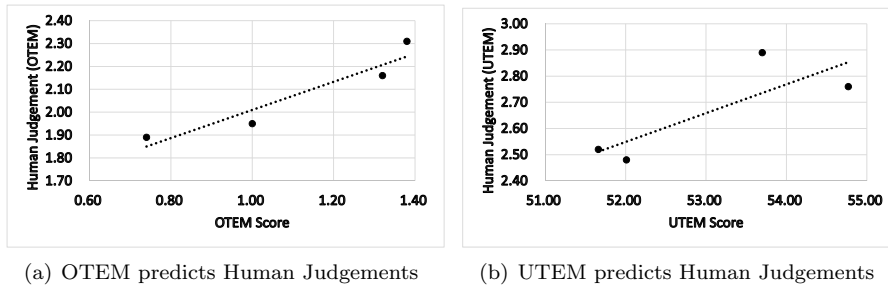

(a) OTEM predicts Human Judgements

(b) UTEM predicts Human Judgements

Figure 3: Correlation between human judgment and OTEM, UTEM. Clear positive correlation is observed for both metrics.

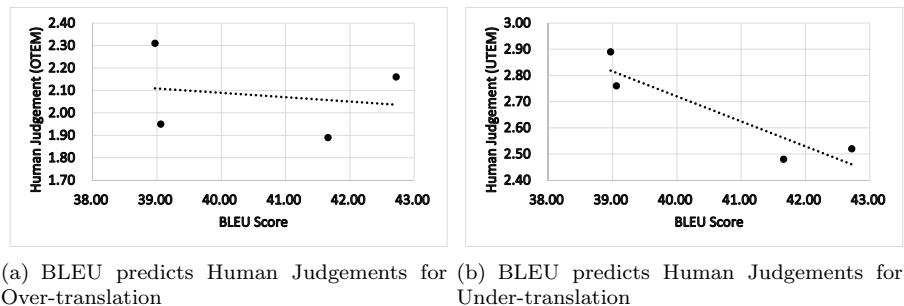

(a) BLEU predicts Human Judgements for Over-translation

(b) BLEU predicts Human Judgements for Under-translation

Figure 4: Correlation between human judgment and BLEU.

been over-translated) for a long sentence with the length of 60 words and a short sentence with the length of 10 words.

### 4.3.2 Correlation with Human Evaluation

We collected the annotated sentence pairs for each NMT system, and summarized the average manually annotated score with the corresponding OTEM and UTEM in Figure 3. We find that both OTEM and UTEM are positively correlated with the manually annotated score. To further verify this observation, we computed the Pearson's r for both metrics, where the value is 0.9461 and 0.8208 for OTEM ($p < 0.05$) and UTEM ($p < 0.05$), respectively. These Pearson's r values strongly suggest that our proposed metrics are indeed highly consistent with human judgment (notice that lower OTEM and UTEM score indicates a better translation).

We also provide comparison between manually annotated score and BLEU score in Figure 4. Obviously, BLEU score demonstrates rather weak association with the over-translation. By contrast, its correlation with the under-translation is much stronger. We conjecture that this is because some important clauses are left untranslated, leading to the occurrence of under-translation, and in con-sequence, the generated translation usually suffers from unfaithfulness issue, a critical aspect for BLEU evaluation. In addition, we also calculated the corresponding Pearson's r between the manually annotated scores and BLEU. The value of Pearson's r for the over- and under-translation is -0.1889 and -0.9192, of which $p$ values are larger than 0.05, indicating that the negative correlation is not significant. In other words, BLEU score is incapable of fully reflecting the over- and under-translation issues.

| Model | Dev | MT02 | MT03 | MT04 |
|---|---|---|---|---|
| PbSMT | 33.09/1.00/56.41 | 34.50/0.84/52.96 | 33.46/0.80/55.46 | 35.23/0.98/55.36 |
| HierSMT | 34.18/**0.78**/**55.77** | 36.21/**0.66**/**52.14** | 34.44/**0.58**/54.88 | 36.91/**0.74**/**54.62** |
| RNNSearch | 34.72/2.05/60.31 | 37.95/1.67/54.68 | 35.23/2.08/56.88 | 37.32/1.78/58.49 |
| Coverage | 35.02/1.30/61.06 | 38.40/1.01/55.09 | 36.18/1.48/55.96 | 37.92/1.27/57.90 |
| FairSeq | 38.84/1.84/58.65 | **41.90**/1.24/53.79 | **40.67**/1.81/54.71 | 42.32/1.89/55.57 |
| Transformer | **38.90**/1.02/57.76 | 41.33/0.77/52.79 | 40.62/0.94/**54.26** | **42.74**/0.83/55.56 |

Table 1: Case-insensitive BLEU-4 / OTEM-2 / UTEM-4 score on NIST Chinese-English translation task. **Bold** highlights the best result among all systems.

## 4.4 Analysis on MT Systems

We summarize the BLEU, OTEM, UTEM scores for different MT systems in Table 1. Particularly, we show OTEM-2(2-gram) rather than OTEM-4(4-gram) because of data sparsity issue.

From Table 1, although all NMT systems outperform all SMT systems with respect to the BLEU score, we observe that for OTEM score, almost all SMT systems outperform all NMT systems. We contribute this to the hard-constraint coverage mechanism in SMT which disables the decoder to repeatedly translate the same source phrases. Sharing similar strength with the coverage mechanism in SMT, Coverage yields substantial improvements over the RNNSearch. It is very interesting that although FairSeq and Transformer produce very similar BLEU scores, Transformer achieves significantly better OTEM scores than FairSeq. We argue that this is because attention in Transformer builds up strong dependencies with both source and previous target words, while convolution in FairSeq can only capture local dependencies.

We can also discover that in terms of UTEM score, all MT systems show similar results, although NMT systems remarkably outperform SMT systems regarding the BLEU score. Through the coverage mechanism, SMT can successfully enforce the translation of each source word. On the contrary, Coverage fails to share this strength. The underlying reason is complicated, which, we argue, requires much more efforts.

Overall, different MT systems show different characteristics with respect to over- and under-translation. BLEU score itself can hardly reflect all these above observations, which highlights the necessity of our work.

## 5   Conclusion

In this paper, we have proposed two novel evaluation metrics, OTEM and UTEM, to evaluate the performance of NMT systems in dealing with over- and under-translation issues, respectively. Although our proposed metrics are based on lexical matching, they are highly correlated to human evaluation, and very effective in detecting the over- and under-translation occurring in the translations produced by NMT systems. Moreover, experimental results show that the coverage mechanism, CNN-based FairSeq and attention-based Transformer possess specific architectural advantages on overcoming these undesired defects.

In the future, we will focus on how to introduce the syntax and semantic similarity between candidates and references to refine our metrics. Besides, inspired by [Shen *et al.*, 2016], we would like to optimize NMT systems directly toward better OTEM and UTEM scores.

## 6   Acknowlwdge

The authors were supported by Natural Science Foundation of China (No. 61672440), the Fundamental Research Funds for the Central Universities (Grant No. ZK1024), Scientific Research Project of National Language Committee of China (Grant No. YB135-49), and Research Fund of the Provincial Key Laboratory for Computer Information Processing Technology in Soochow University (Grant No. KJS1520). We also thank the reviewers for their insightful comments.